%% file: main.tex
\documentclass[10pt,twocolumn,letterpaper]{article}

\usepackage{cvpr}              

\usepackage{graphicx}
\usepackage{amsmath}
\usepackage{amssymb}
\usepackage{booktabs}
\usepackage{lipsum}
\usepackage{booktabs}
\newcommand\blfootnote[1]{%
  \begingroup
  \renewcommand\thefootnote{}\footnote{#1}%
  \addtocounter{footnote}{-1}%
  \endgroup
}
\usepackage[accsupp]{axessibility}

\usepackage[pagebackref,breaklinks,colorlinks]{hyperref}

\usepackage[capitalize]{cleveref}
\crefname{section}{Sec.}{Secs.}
\Crefname{section}{Section}{Sections}
\Crefname{table}{Table}{Tables}
\crefname{table}{Tab.}{Tabs.}

\def\cvprPaperID{98} 
\def\confName{CVPR}
\def\confYear{2022}

\begin{document}

\title{FS-NCSR: Increasing Diversity of the Super-Resolution Space via \\ Frequency Separation and Noise-Conditioned Normalizing Flow}

\author{Ki-Ung~Song \textsuperscript{\rm 1, 3}\thanks{Equal contribution}
\quad~Dongseok~Shim\textsuperscript{\rm 1, 3}\footnotemark[1]
\quad~Kang-wook~Kim\textsuperscript{\rm 1, 3}\footnotemark[1]
\\ ~Jae-young~Lee\textsuperscript{\rm 1, 3}
\quad~Younggeun~Kim\textsuperscript{\rm 1, 2, 3}\\
		{\textsuperscript{\rm 1} Seoul National University, Seoul, Republic of Korea
		}  \\
		{\textsuperscript{\rm 2}MINDsLab Inc., Gyeonggi, Republic of Korea} \\
		{\textsuperscript{\rm 3}Deepest, Seoul, Republic of Korea } \\
        \small{\texttt{\{sk851, tlaehdtjr01, full324, jerry96, eyfydsyd97\}@snu.ac.kr}}
	}
\maketitle

\begin{abstract}

Super-resolution suffers from an innate ill-posed problem that a single low-resolution (LR) image can be from multiple high-resolution (HR) images. Recent studies on the flow-based algorithm solve this ill-posedness by learning the super-resolution space and predicting diverse HR outputs. Unfortunately, the diversity of the super-resolution outputs is still unsatisfactory, and the outputs from the flow-based model usually suffer from undesired artifacts which causes low-quality outputs. In this paper, we propose FS-NCSR which produces diverse and high-quality super-resolution outputs using frequency separation and noise conditioning compared to the existing flow-based approaches.  As the sharpness and high-quality detail of the image rely on its high-frequency information, FS-NCSR only estimates the high-frequency information of the high-resolution outputs without redundant low-frequency components. Through this, FS-NCSR significantly improves the diversity score without significant image quality degradation compared to the NCSR, the winner of the previous NTIRE 2021 challenge.
\end{abstract}
\blfootnote{This work was done as a project of Deepest}

\section{Introduction} \label{sec:intro}
\input{paper/1_intro}

\section{Related Works} \label{sec:related}
\input{paper/2_related}

\section{Methods}
\input{paper/3_methods}

\section{Experiments}
\input{paper/4_experiments}

\section{NTIRE 2022 Challenge}
\input{paper/5_ntire}

\section{Conclusion}
\input{paper/6_conclusion}

\section{Acknowledgement}
This work was supported by Institute of Information \& communications Technology Planning \& Evaluation (IITP) grant funded by the Korea government(MSIT) [NO.2021-0-01343, Artificial Intelligence Graduate School Program (Seoul National University)]

{
\newpage
\small
\bibliographystyle{ieee_fullname}
\bibliography{egbib}
}

\end{document}

%% file: paper/1_intro.tex
Single image super-resolution (SISR), the task that restores low-resolution (LR) images to high-resolution (HR) images, is an active research topic that can be utilized in several applications such as surveillance \cite{zou2011very}, medical and astronomical image processing \cite{shi2013cardiac, chen2018efficient, li2018super}.

Early SISR approaches \cite{wang2014srcnn, kim2016accurate, lim2017enhanced, zhang2018image, zhang2018residual} focus on generating a single high-quality output for a given input LR image by improving Peak Signal-to-Noise Ratio (PSNR) ratio between the input LR images and predicted HR outputs.
Since those studies utilize $L_{1}$ or $L_{2}$ loss between the generated and ground-truth HR images, they suffer from an over-smoothing problem.
Alternative to PSNR-oriented models, GAN-based methods \cite{ledig2017srgan, wang2018esrgan} are proposed to generate photo-realistic super-resolved images.

Unfortunately, multiple possible HR images exist for a single LR image and the aforementioned deterministic models which improve the image quality of a single output cannot solve this ill-posed nature of the super resolution.
SRFlow \cite{lugmayr2020srflow} learns the distribution of the HR image consistent for the given LR images and predicts diverse HR images to improve the high photo-realism, diversity, and the LR consistency at once.
Following, NCSR \cite{kim2021noise} adopts noise-conditioned layers suggested in SoftFlow \cite{kim2020softflow} and HCFlow \cite{liang21hierarchical} proposes hierarchical conditional flow for the diversity and the higher image quality.
However, flow-based models usually generate undesired artifacts in HR outputs which leads to lower image quality and the diversity of the outputs are not improved significantly compared to SRFlow.

We observe that the super-resolution models predict the missing high-frequency information of the HR images from the given LR image which takes part in generating the diverse details of the HR images such as the shape of the foliage and the direction of the fur.
Previous super-resolution models \cite{lugmayr2020srflow,kim2021noise} predict not only high-frequency information, but also low-frequency information of the HR images.
It leads to inefficient training and these models have difficulty in increasing the diversity and the image quality of the super-resolution outputs.

In this paper, we propose \textbf{FS-NCSR} (Frequency-Separated Noise-Conditioned Normalizing Flow for Super-Resolution) which applies frequency separation to NCSR.
We reconstruct the low-frequency information of the HR outputs by upsampling LR images in bicubic without any learnable parameters and predict the high-frequency information by training flow-based model.
By doing so, we increase the diversity of learned super-resolution space in both $\times$4 and $\times$8 settings and improve the super-resolution quality by reducing the number of the artifact. Our contributions can be summarized as follows:
\begin{itemize}
    \item We propose a flow-based algorithm for high-quality diverse super-resolution output using noise-conditioned affine coupling and frequency separation.
    \item By filtering low-resolution information of the target image, the generative model focuses on producing high-frequency outputs and improves super-resolution quality.
    \item We expand the filtered input data distribution by adding noise to the sparse high-frequency image for the output diversity.
\end{itemize}

%% file: paper/2_related.tex
\subsection{Single Image Super Resolution}
Super-resolution has been studied long in computer vision fields. Before deep learning-based methods have been applied, sparsed coding\cite{dai2015jointlyRegressedRegressors, sun2012sceneMatching, yang2008sparseRepr, yang2010sparseRepr} and local linear regression\cite{timofte2013anchNeighReg, timofte2014A+, yang2013simpleFuncSR} have been highly applied. Many deep learning-based methods have been approached for SISR, since SRCNN \cite{wang2014srcnn} which exploited CNN layers and L1 Loss. After SRCNN was proposed, many variations have been suggested including \cite{wang2018esrgan}. But as CNN-based methods have relied on L1 or L2 loss, they have generated blurry images. GAN-based methods, which were first suggested by SRGAN\cite{ledig2017srgan}, have shown improvements by employing adversarial loss and perceptual loss. Although GAN-based methods have generated images with good quality\cite{ledig2017srgan, wang2018esrgan}, their diversities were so limited, thereby generating only one image.

\subsection{Normalizing Flow}
Flow-based models have been first proposed by \cite{dinh2014nice} for modeling complex high dimensional density. As flow-based models learn the whole distribution, they have been widely used for mapping complex distributions given simple distribution, including Gaussian distribution. Invertible neural networks have been adopted to map complex distributions from simple distributions\cite{dinh2014nice, dinh2017realnvp, kingma2018glow}. Flow-based models in the early days have not shown great improvements relative to GAN-based models. However, SRFlow\cite{lugmayr2020srflow}, which adopts negative log-likelihood loss, showed improvements in image quality and diversities simultaneously. As SRFlow used negative log-likelihood loss, it could learn the whole distribution, which leads to generating much more diverse images than GAN-based methods. NCSR \cite{kim2021noise} has shown further improvements in terms of image quality and diversity, by providing networks with noises. \cite{kim2021noise} has proposed adding a conditional noise layer, which essentially resolves distribution discrepancy between simple data and complex data.

\subsection{Frequency Separation}
The study of frequency domain based on Fast Fourier Transform (FFT) algorithm \cite{cooley1965algorithm} played a crucial role in traditional signal processing. In this perspective, before the era of deep learning, studying the frequency information was important in image restoration research. In this light, it is readily known that high-frequency information of the given image contributes greatly to its sharpness and high-quality detail. Therefore, we can say that a recent huge success of deep learning-based approaches in realistic images generation is due to the success of synthesizing high-frequency information of the desired images.

Therefore, in recent image restoration research including super-resolution, there exist approaches \cite{whang2021deblurring} in which low-frequency and high-frequency are separated and treated by a separate neural network, and approaches \cite{suvorov2021resolution} in which an FFT-based layer is designed to better process information of the frequency domain. 
We observed that when the former approaches were combined with NCSR, instability of the NLL training of the flow-based model occurred. 
And in the case of the latter approaches, the existing FFT-based layers are not suitable for the flow-based approach due to their non-invertible nature.

%% file: paper/3_methods.tex
Given a LR image, our goal is to learn a diverse super-resolution space corresponding to that image. From the perspective of the frequency domain, we propose a more efficient method to increase the diversity of learned space. In this section, we introduce our point of view and proposed method. We begin with a brief background related to our work.

\begin{figure*}
\begin{center}
\includegraphics[width=0.9\linewidth]{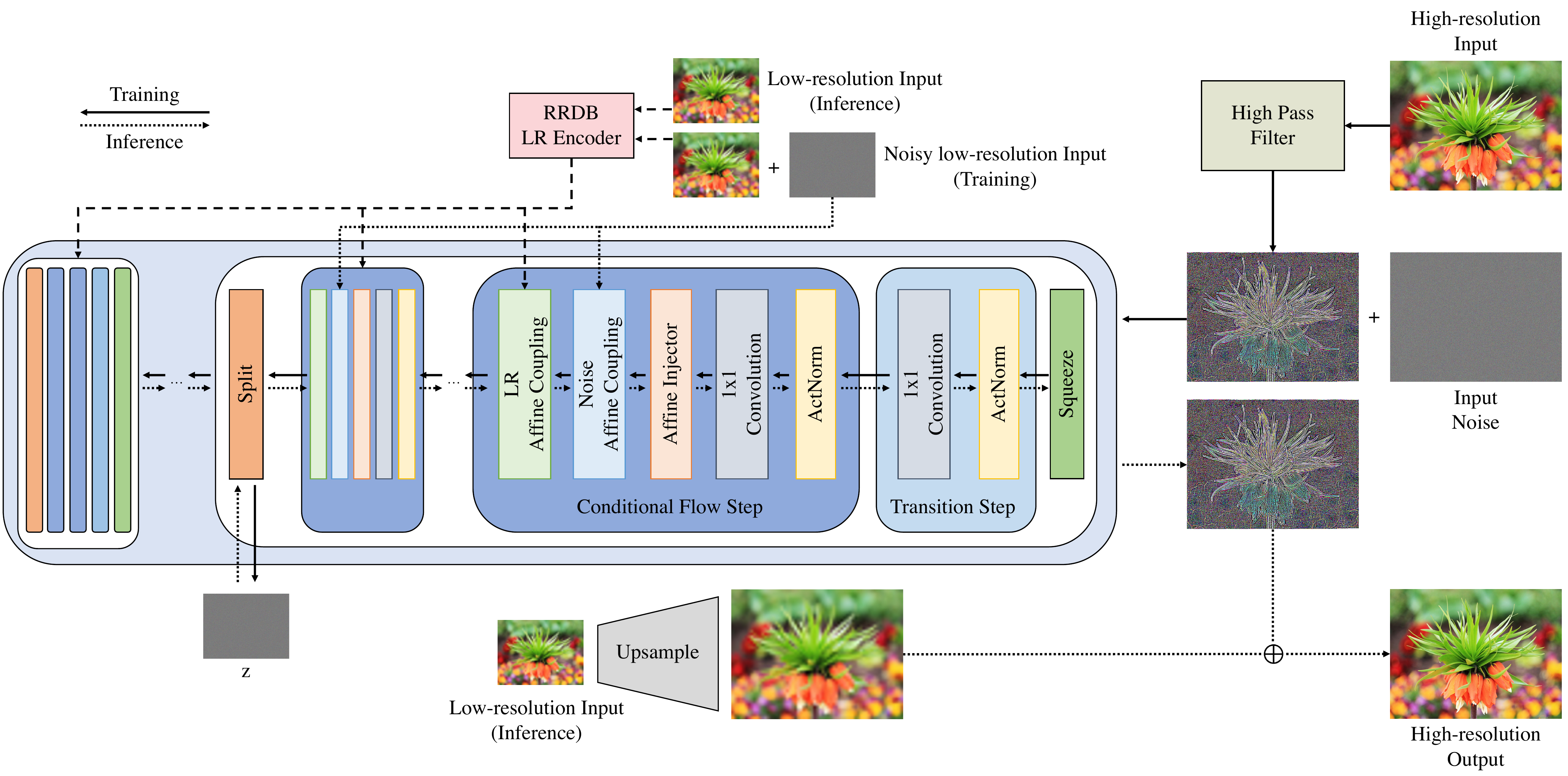}
\end{center}
\caption{Algorithm overview. We propose a frequency separation on the target image and applies noise on high-frequency input with noise-conditioned coupling layers for diverse super-resolution outputs.}
\label{fig:overall}
\end{figure*}

\subsection{Background}
Various model frameworks (\eg Generative Adversarial Networks \cite{goodfellow2014generative}, Normalizing Flow \cite{rezende2015variational}, and Diffusion probabilistic models \cite{ho2020denoising}) have been proposed in recent deep learning-based generative model research. And they show their respective strengths and weaknesses along with excellent performance.
Among them, the flow-based model configures a mapping $f_{\theta}:X\rightarrow Z$ between the desired data distribution $X$ and latent space distribution $Z$ (\eg Gaussian) through a series of invertible transformations. Such an invertible mapping architecture enables an explicit computation of negative log-likelihood (NLL) by the change of variable formula as:
\begin{equation}\label{eq:eq_1}
    -\log{p_{X}(x)} = -\log{p_{Z}(f_{\theta}(x))}-\log{|\det{\frac{\partial{f_\theta}}{\partial{x}}}(x)|}.
\end{equation}
By minimizing NLL directly, it is widely known that the flow-based models show decent performance in mode coverage of the desired data distribution.

Based on this advantage of the flow-based approach, SRFlow \cite{lugmayr2020srflow} first showed that the flow-based modeling of the conditional distribution of the HR image can successfully learn super-resolution space corresponding to the given LR input. And one of its variants model, NCSR \cite{kim2021noise}, proposed an additional noise-conditional layer to SRFlow to generate more diverse super-resolution outputs.
Results of the previous works show that the ill-posedness of super-resolution can be solved from the perspective of super-resolution space learning. To take advantage of the flow-based model's good mode coverage performance, we propose a method to learn more diverse super-resolution space with NCSR architecture.

\subsection{High-Frequency Information}
There are various ways to configure a High-pass filter and Low-pass filter to separate high-frequency and low-frequency information. Without affecting the stability of NLL training of the flow-based model, we utilize the bicubic downsampling-upsampling process as the Low-pass filter, $L_{s}$, with a specific scale factor $s$. And the corresponding High-pass filter, $H_s$, computes the high-frequency information $x_{hf}$ of the given input by subtracting low-frequency information from the HR target $x$:
\begin{equation}\label{eq:eq_2}
    L_{s}(x) = ((x)_{s\downarrow})_{s\uparrow}, \quad
    H_{s}(x) = x_{hf} = x - L_{s}(x).
\end{equation}

There are also other frequency separation methods. Some can configure $L_{s}$ and $H_s$ based on FFT and others can utilize the known 3x3 (or 5x5) kernel.
In the former case, the filtering threshold level is an additional hyperparameter that is heavily dependent on an individual image. And in both cases, to match the low-frequency information of the LR input $y$ and $L_{s}(x)$, additional process such as the usage of a neural network is required leading to instability of NLL training.

By using this simple kind of High-pass filter, sparse high-frequency information can be efficiently obtained since we have the LR input as $y = (x)_{s\downarrow}$. And it leads to our proposed method which achieves efficient training without the need for additional memory or network compared to the previous flow-based approaches.

\subsection{Overall Method}
We propose FS-NCSR (Frequency Separating Noise-Conditioned Normalizing Flow for Super-Resolution), the generative model for super-resolution only produces the high-frequency information of the target HR image $x$ without redundant low-frequency information readily available from $y = (x)_{s\downarrow}$. Our overall model architecture is shown in Figure \ref{fig:overall}.

In the training process of the flow-based models, dequantization processes exist \cite{kingma2018glow, ho2019flow++} for better performance. As can be readily checked in Figure \ref{fig:gt_freq} and Table \ref{tab:Freq_space}, the high-frequency information is relatively sparse compared to HR images. And training the model with this kind of information is difficult. In the previous work of NCSR, the idea of Softflow \cite{kim2020softflow} was used by adding a different level of noise to the input instead of the dequantization process.
This can be interpreted as an attempt to expand the modality of the desired data distribution's sparse region in the perspective of score matching \cite{song2019generative, song2020score} which is in the spotlight of the generative model today.
Therefore, we applied the same idea of Softflow \cite{kim2020softflow} to deal with sparse information, and it was crucial in the training stability of the proposed method.

Now, with the same analog to the work of \cite{lugmayr2020srflow, kim2021noise}, we can formulate the training process of our method as follows:
\begin{equation}\label{eq:eq_3}
    \begin{aligned}
    x^{+}_{hf} = x_{hf} + v, \quad v \sim \mathcal{N}(0, \Sigma), \\
    y^{+} = y + w, \quad w \sim \mathcal{N}(0, \hat{\Sigma}),\\
    z = f_{\theta}(x^{+}_{hf}|y^{+}, v).
    \end{aligned}
\end{equation}
where $w$ indicates noise resized to the same size as the LR input $y$.
And also similar to \cite{lugmayr2020srflow, kim2021noise}, we formulate the loss function only NLL $\mathcal{L}_{nll}$ as below,
\begin{equation}\label{eq:eq_4}
    \begin{aligned}
    \mathcal{L}_{nll} &= -\log{ p_{X}(x|y^{+},v)} \\
    &= -\log{p_{Z}(f_{\theta}(x;y^{+},v))} - \log{|\det{\frac{\partial{f_\theta}}{\partial{y}}}(x;y^{+},v)|}.
    \end{aligned}
\end{equation}

The model trained in proposed method does not require additional cost in the inference stage compared to the previous approaches. Since the low-frequency information $L_{s}(x)$ is readily given by the LR input $y = (x)_{s\downarrow}$. The super-resolution output $\hat{x}$ is obtained by:
\begin{equation}\label{eq:inference}
\begin{aligned}
    \hat{x} = f^{-1}_{\theta}(z;y,v) + (y)_{s\uparrow} \\
    = f^{-1}_{\theta}(z;(x)_{s\downarrow},v) + L_{s}(x).
\end{aligned}
\end{equation}
where $v$ is the random noise from the latent space $Z$.

In this perspective of frequency domain, super-resolution is the process of generating the corresponding high-frequency information since we have $f^{-1}_{\theta}(\cdot;(x)_{s\downarrow}) \approx H_{s}(x)$.

%% file: paper/4_experiments.tex
\subsection{Datasets}
We utilize DF2K dataset, a merged dataset of DIV2K \cite{agustsson2017ntire} and Flickr2K\footnote{\url{https://github.com/limbee/NTIRE2017}}, for training and evaluation.
DIV2K dataset consists of 800, 100, and 100 high-resolution images of train, validation, and test split, respectively.
Flickr2K dataset comprises 2560 high-resolution images.
The train split from the DIV2K dataset and the whole Flickr2K dataset are merged and used for training.
We evaluate our model with the validation split of DIV2K dataset.

We try to increase the amount of training dataset by including crawled images from Unsplash website\footnote{\url{https://unsplash.com}}, but there are no performance improvements in the diversity and visual quality of super-resolved images.
Thus, we do not include our crawled dataset in this research.

During training, we randomly crop 160×160 patches from original HR images and use them as HR samples.
We obtain LR samples by downsampling HR patches and utilizing HR and LR patches as HR-LR pairs for training.
The LR samples are downsampled via bicubic kernel.
We train our model in RGB channels, and randomly apply horizontal flips and 90-degree rotation for data augmentation.
\begin{figure}[t]
    \centering
    \includegraphics[width=1.0\linewidth]{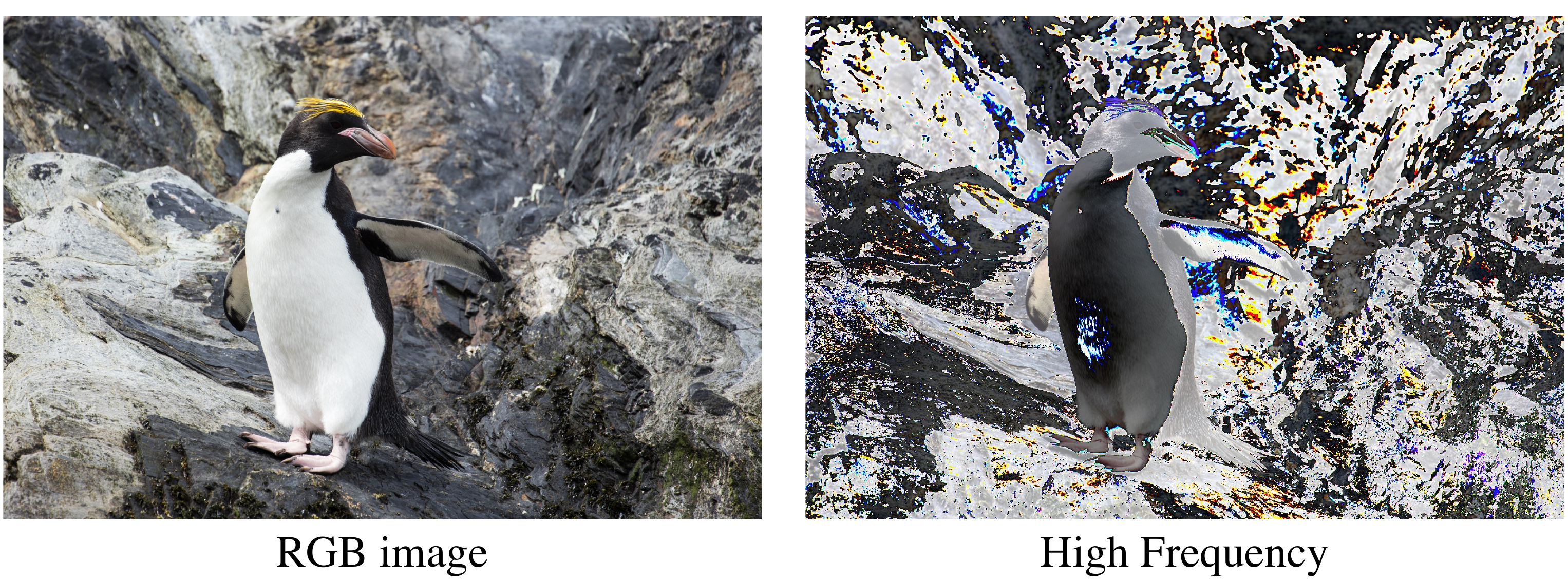}
    \caption{Full RGB vs. High-frequency information. High-frequency information is relatively sparse compared to its original RGB images. The high-frequency information was obtained based on $\times$128 scale for clear visual confirmation. For $\times$4 and $\times$8 cases, the high-frequency information is much sparser, making it difficult to see.}
    \label{fig:gt_freq}
\end{figure}

\subsection{Training}

We use the Adam optimizer \cite{kingma2014adam} with $\beta_1$ = 0.9, $\beta_2$ = 0.99, $\epsilon$ = $10 ^{-8}$, and set the initial learning rate as $2 \times 10 ^{-4}$.
Following \cite{kim2021noise}, the learning rate is halved at 50\%, 75\%, 90\%, and 95\% of the total training steps.
We train our network with a batch size of 16 on a V100 GPU.
The $\times$4 network was trained at 180k steps and the $\times$8 network at 220k steps.

\subsection{Evaluation}
We evaluate our model and other baselines based on three criteria: photo-realism, diversity of super-resolution space, and image consistency on LR.
We adopt LPIPS \cite{zhang2018unreasonable} to evaluate photo-realism, diversity score to evaluate diversity, and LR PSNR to evaluate LR consistency.

\noindent\textbf{LPIPS.}
LPIPS is the distance between the super-resolved and the ground-truth HR image.
The distance is measured on the feature space of AlexNet \cite{krizhevsky2012imagenet}.

\noindent\textbf{Diversity Score.}
To obtain meaningful diversity of models, Lugmayr \etal{} \cite{Lugmayr_2021_CVPR} proposed the diversity score.
Let the ground-truth HR image $y$ and $y_{k}$ be the $k$-th patch of $y$.
Generating $M$ samples from the super-resolution models, the $i$-th super-resolved images from the model is $\hat{y^{i}}$, and its $k$-th patch is $\hat{y_{k}^{i}}$, where $i \in \{1, 2, ..., M\}$.
Than the diversity score $S_{M}$ can be computed as follows:
\begin{equation}\label{eq:div_score}
    S_{M} = \cfrac{1}{\bar{d_{M}}} \left(\bar{d_{M}} - \cfrac{1}{K} \sum_{k=1}^{K}\min{\left\{d\left(y_{k}, \hat{y_{k}^{i}}\right)\right\}}^{M}_{i=1} \right),
\end{equation}
where minimum distance on a global sample, $\bar{d_{M}}$, defined as follows:
\begin{equation}\label{eq:avg_distance}
    \bar{d_{M}}=\min{\left\{ \cfrac{1}{K} \sum_{k=1}^{K}d\left(y_{k}, \hat{y_{k}^{i}}\right) \right\}}^{M}_{i=1}.
\end{equation}
We use LPIPS as distance function $d$, and set $M=10$.

\noindent\textbf{LR PSNR.}
In LR, the super-resolved output of the model must be consistent with the original LR input.
Thus, we measure PSNR (Peak Signal-to-Noise Ratio) between downsampled super-resolved image and given input LR image.

\begin{table}[t]
\begin{tabular}{@{}llll@{}}
\toprule
Model & Diversity$\uparrow$ & LPIPS$\downarrow$ & LR PSNR$\uparrow$ \\ \midrule
RRDB~\cite{wang2018esrgan}  & 0 & 0.253 & 49.20   \\
ESRGAN~\cite{wang2018esrgan}  & 0 & 0.124 & 39.03   \\
ESRGAN+~\cite{Rakotonirina_2020} & 22.13 & 0.279 & 35.45   \\
SRFlow~\cite{lugmayr2020srflow} & 25.26 & 0.120 & 49.97   \\
HCFlow~\cite{liang21hierarchical} & 22.73 & \textbf{0.116} & 49.46 \\
NCSR~\cite{kim2021noise}  & 26.72 & 0.119 & \textbf{50.75}   \\
\textbf{FS-NCSR (Ours)} & \textbf{29.44} & 0.127 & 49.31  \\ \bottomrule
\end{tabular}
\captionsetup{justification=raggedright,singlelinecheck=false}
\vspace{-0.3cm}
\caption{General image $\times$4 super-resolution results on the DIV2K validation set.  We measure all the metrics with $M=10$ samples for each HR image.}
\label{tab:X4_results}
\end{table}

\subsection{Quantitative Results} \label{sub:quantitative}
We compare our model, FS-NCSR, with diverse baseline models: RRDB \cite{wang2018esrgan}, ESRGAN \cite{wang2018esrgan}, ESRGAN+ \cite{Rakotonirina_2020}, SRFlow \cite{Lugmayr_2021_CVPR}, HCFlow \cite{liang21hierarchical}, and NCSR \cite{kim2021noise}.
RRDB is the model trained with $L_{1}$ loss with ground-truth HR image, consequently oriented to minimizing PSNR.
ESRGAN and ESRGAN+ are GAN-based methods that are the common baselines for photo-realistic super-resolution.
RRDB and ESRGAN are deterministic models, so their diversity scores are zero.
SRFlow, HCFlow, and NCSR are stochastic super-resolution models that can super-resolve diverse photo-realistic images from the given input LR image.
For all the flow-based super-resolution models, the temperature is set to 0.9. However, the temperature is 0.85 for NCSR $\times$8 model.

We measure the diversity score, LPIPS, LR PSNR of our model and compare them with the reported results of other baselines.
We evaluate all the models in $\times$4 super-resolution setting.
As shown in Table \ref{tab:X4_results}, our proposed model, FS-NCSR, achieves the highest diversity score in $\times$4 setting.
The diversity score of FS-NCSR is significantly higher than NCSR \cite{kim2021noise}, which indicates frequency separation plays a key role to improve diversity.
Although FS-NCSR achieves the lower LR PSNR and higher LPIPS than SRFlow \cite{lugmayr2020srflow}, HCFlow \cite{liang21hierarchical} and NCSR, diversity increase is significant compared to such performance degradation so can be compensated.
In addition, we observe that the number of artifacts and failure cases in the generated samples of FS-NCSR is less than that of NCSR.
We will discuss this qualitative comparison in \ref{sub:qualitative}.

We also evaluate all the models except ESRGAN+ \cite{Rakotonirina_2020} in $\times$8 super-resolution setting.
As presented in Table \ref{tab:X8_results}, FS-NCSR outperforms all the other methods in terms of diversity score and LPIPS.
Also, FS-NCSR achieves comparable LR PSNR with SRFlow \cite{lugmayr2020srflow}, the model which achieved the highest LR PSNR.
These results show that FS-NCSR outperforms all the other methods in terms of photo-realism and diversity, and frequency separation is a decisive factor.

\begin{table}[t]
\begin{tabular}{@{}llll@{}}
\toprule
Model & Diversity$\uparrow$ & LPIPS$\downarrow$ & LR PSNR$\uparrow$ \\ \midrule
RRDB~\cite{wang2018esrgan} & 0 & 0.419 & 45.43   \\
ESRGAN~\cite{wang2018esrgan} & 0 & 0.277 & 31.35   \\
SRFlow~\cite{lugmayr2020srflow} & 25.31 & 0.272 & \textbf{50.00}   \\
NCSR~\cite{kim2021noise} & 26.8 & 0.278 & 44.55   \\
\textbf{FS-NCSR (Ours)} & \textbf{26.9} & \textbf{0.257} & 48.90 \\ \bottomrule
\end{tabular}
\captionsetup{justification=raggedright,singlelinecheck=false}
\vspace{-0.3cm}
\caption{General image $\times$8 super-resolution results on the DIV2K validation set. We measure all the metrics with $M=10$ samples for each HR image.}
\label{tab:X8_results}
\end{table}

\begin{figure}[t]
    \centering
    \includegraphics[width=0.8\linewidth]{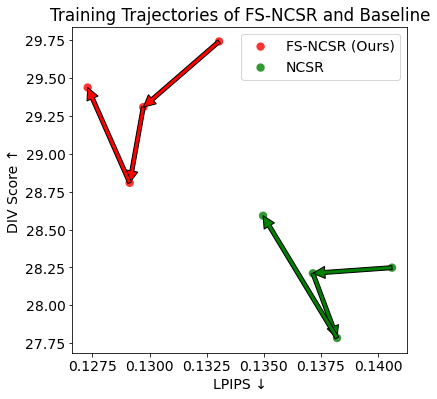}
    \vspace{-1.\baselineskip}
    \caption{%
    LPIPS and diversity scores of multiple checkpoints of FS-NCSR and NCSR \cite{kim2021noise}. The super-resolution ratio is $\times$4 setting. We measure LPIPS and diversity at 150k, 160k, 170k, 180k steps of training procedures. The arrows in the figure indicate the change in both metrics as the iteration increases by 10k.
    }
    \label{fig:trajectory}
\end{figure}

To clearly demonstrate the effect of frequency separation, we additionally report the metric trajectories during the training process of FS-NCSR and NCSR \cite{kim2021noise}.
We measure LPIPS and diversity score in 150k, 160k, 170k, 180k steps for each model.
The results of such models during the training process are presented in Figure \ref{fig:trajectory}.
For trained weights of FS-NCSR, higher diversity and lower LPIPS than NCSR weights of the same iteration are measured.
These results show that frequency separation consistently improves the diversity and photo-realism of the model output during the training process.

\begin{figure*}
\begin{center}
\centering
    \begin{subfigure}{.24\textwidth}
    \centering
    \includegraphics[width=\linewidth]{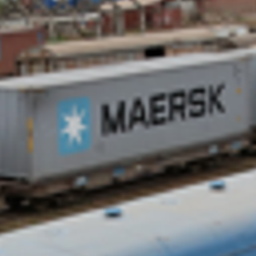}
    \caption{$\times$4 LR}
    \end{subfigure}
    \begin{subfigure}{.24\textwidth}
    \centering
    \includegraphics[width=\linewidth]{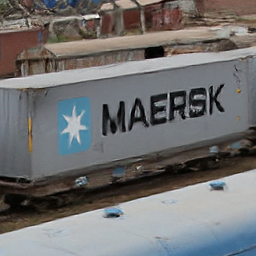}
    \caption{NCSR \cite{kim2021noise}}
    \end{subfigure}
    \begin{subfigure}{.24\textwidth}
    \centering
    \includegraphics[width=\linewidth]{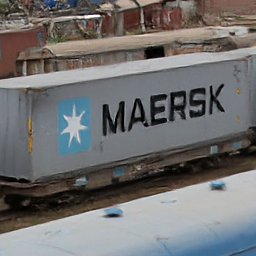}
    \caption{FS-NCSR (Ours)}
    \end{subfigure}
    \begin{subfigure}{.24\textwidth}
    \centering
    \includegraphics[width=\linewidth]{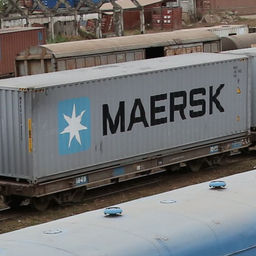}
    \caption{Ground Truth}
    \end{subfigure}
    \vspace{-0.3cm}
\caption{Qualitative results with comparison to NCSR on the DIV2K validation set for SR $\times$4 results. The cropped part of the ground truth is from 0850 from DIV2K. Each output of NCSR and FS-NCSR was chosen randomly from 10 generated outputs respectively.}
\label{fig:compare_crop}
\end{center}
\end{figure*}

\begin{figure*}
\begin{center}
\centering
    \begin{subfigure}{.16\textwidth}
    \centering
    \includegraphics[width=\linewidth]{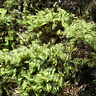} 
    \caption{Output 1}
    \end{subfigure}
    \begin{subfigure}{.16\textwidth}
    \centering
    \includegraphics[width=\linewidth]{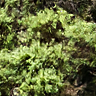}
    \caption{Output 2}
    \end{subfigure}
    \begin{subfigure}{.16\textwidth}
    \centering
    \includegraphics[width=\linewidth]{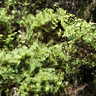}
    \caption{Output 3}
    \end{subfigure}
    \begin{subfigure}{.16\textwidth}
    \centering
    \includegraphics[width=\linewidth]{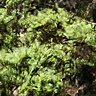}
    \caption{Output 4}
    \end{subfigure}
    \begin{subfigure}{.16\textwidth}
    \centering
    \includegraphics[width=\linewidth]{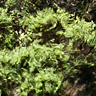}
    \caption{Output 5}
    \end{subfigure}
    \begin{subfigure}{.16\textwidth}
    \centering
    \includegraphics[width=\linewidth]{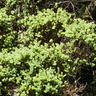}
    \caption{Ground Truth}
    \end{subfigure}
    \vspace{-0.3cm}
\caption{Qualitative result to check the FS-NCSR's diversity of generated details on the DIV2K validation set for $\times$4 super-resolved results. The ground truth is a cropped part of 0875 from DIV2K. 5 outputs were chosen from 10 generated FS-NCSR outputs.}
\label{fig:diversity}
\end{center}
\end{figure*}

\begin{figure*}
\begin{center}
\centering
    \begin{subfigure}{.45\textwidth}
    \centering
    \includegraphics[width=.95\linewidth]{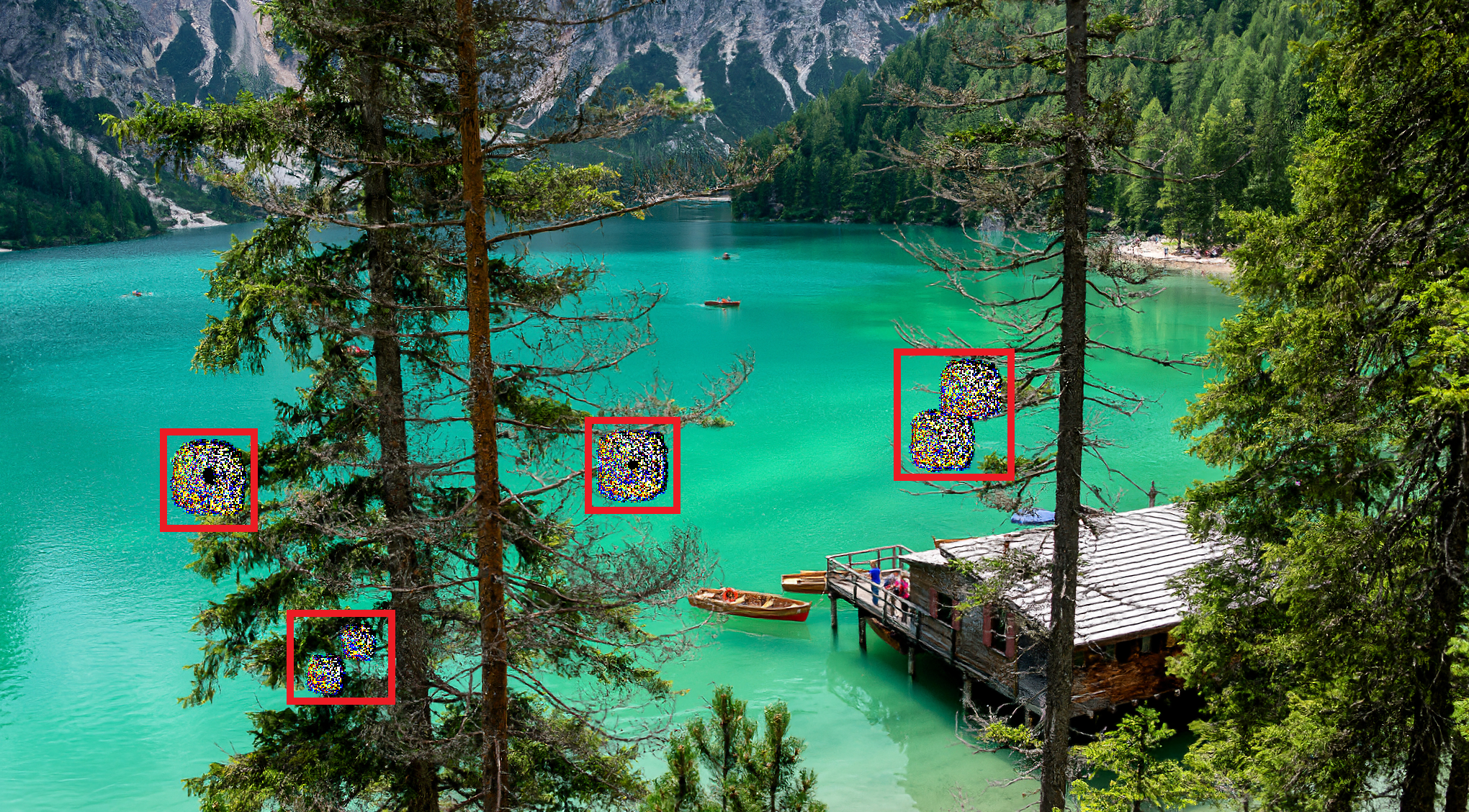}
    \caption{Failure case of NCSR}
    \end{subfigure}
    \begin{subfigure}{.45\textwidth}
    \centering
    \includegraphics[width=.95\linewidth]{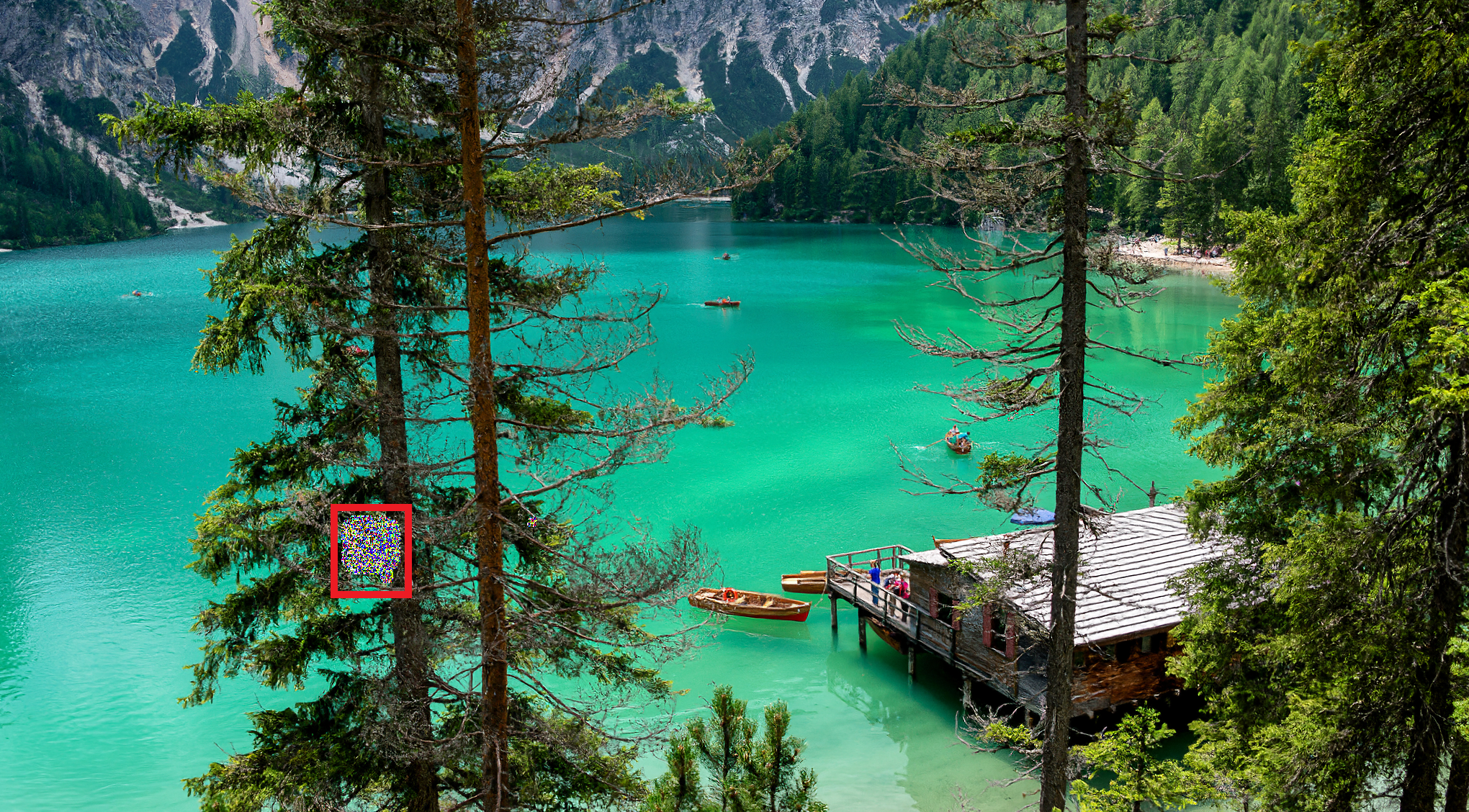}
    \caption{Failure case of FS-NCSR}
    \end{subfigure}
    
    \begin{subfigure}{.45\textwidth}
    \centering
    \includegraphics[width=.95\linewidth]{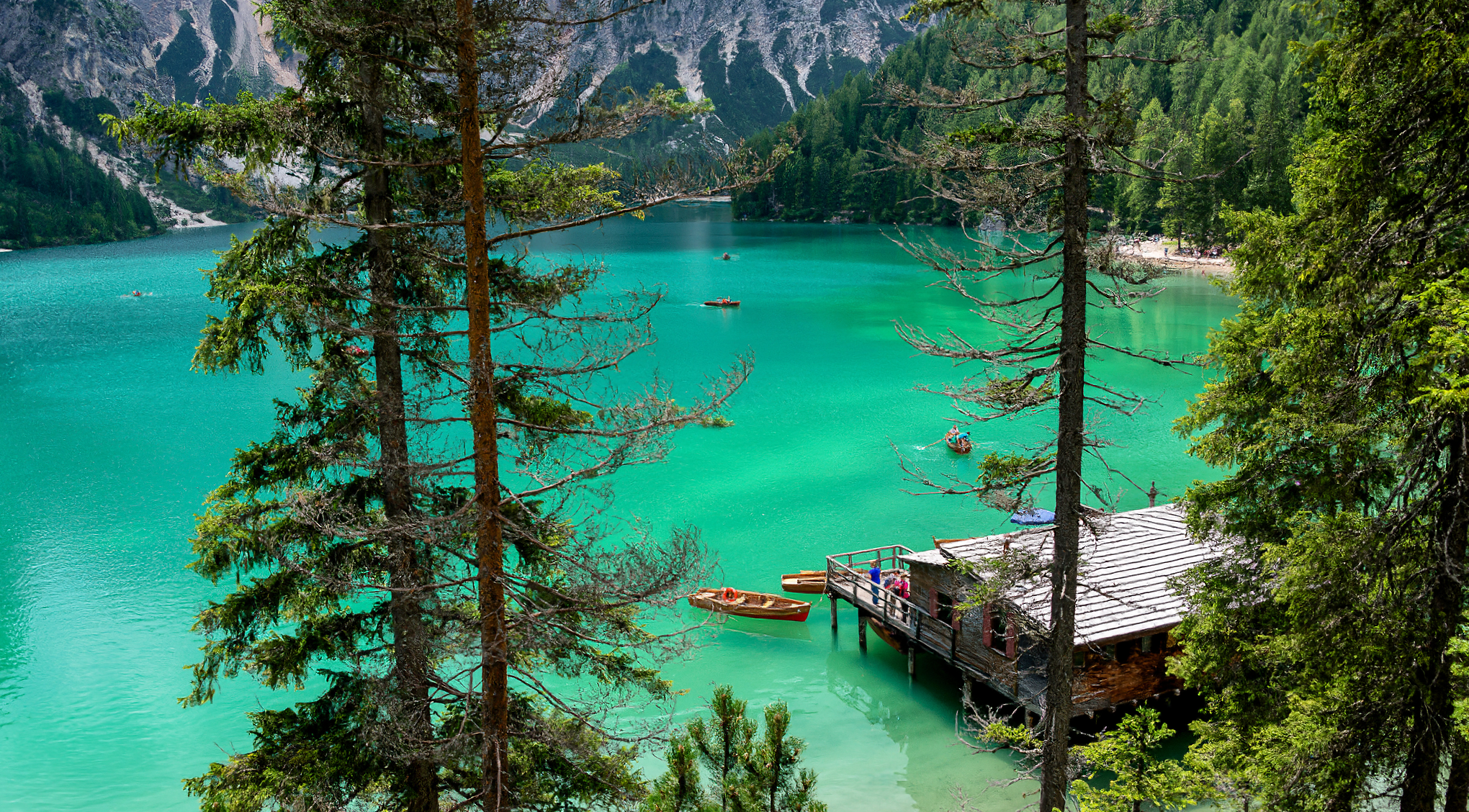}
    \caption{FS-NCSR Ouput without any artifact}
    \end{subfigure}
    \begin{subfigure}{.45\textwidth}
    \centering
    \includegraphics[width=.95\linewidth]{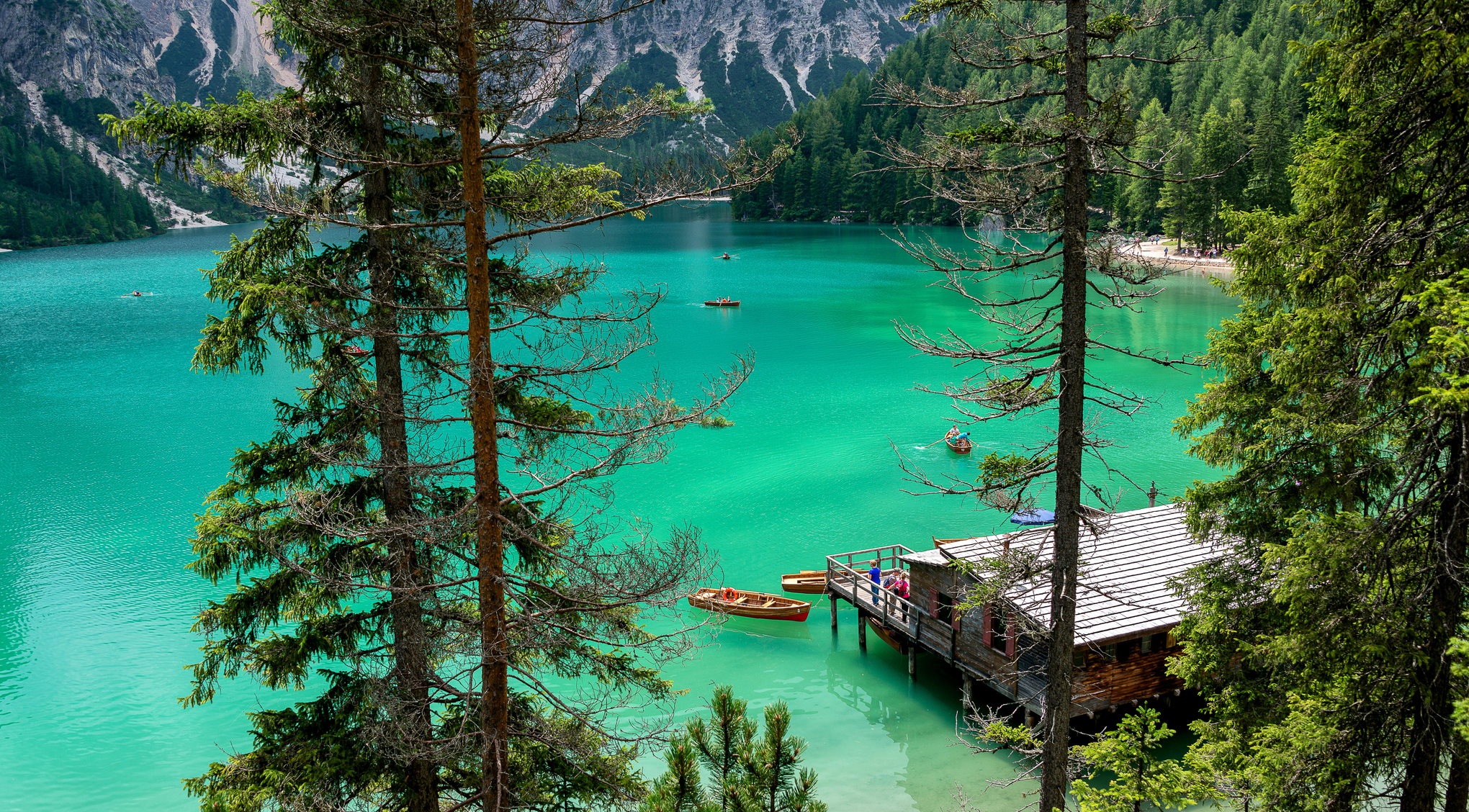}
    \caption{Ground Truth: 0807 from DIV2K}
    \end{subfigure}
    
    \begin{subfigure}{.45\textwidth}
    \centering
    \includegraphics[width=.95\linewidth]{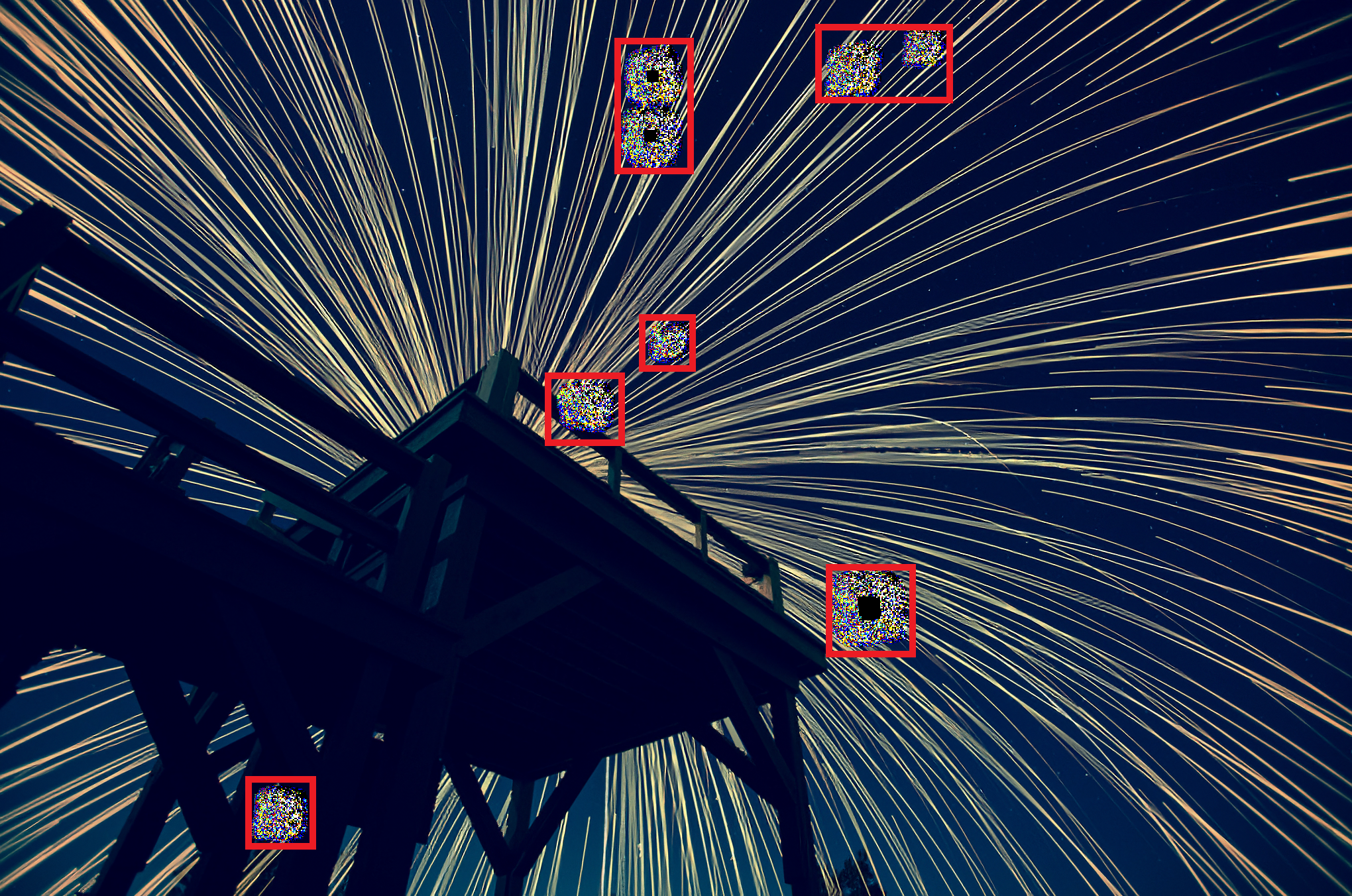}
    \caption{Failure case of NCSR}
    \end{subfigure}
    \begin{subfigure}{.45\textwidth}
    \centering
    \includegraphics[width=.95\linewidth]{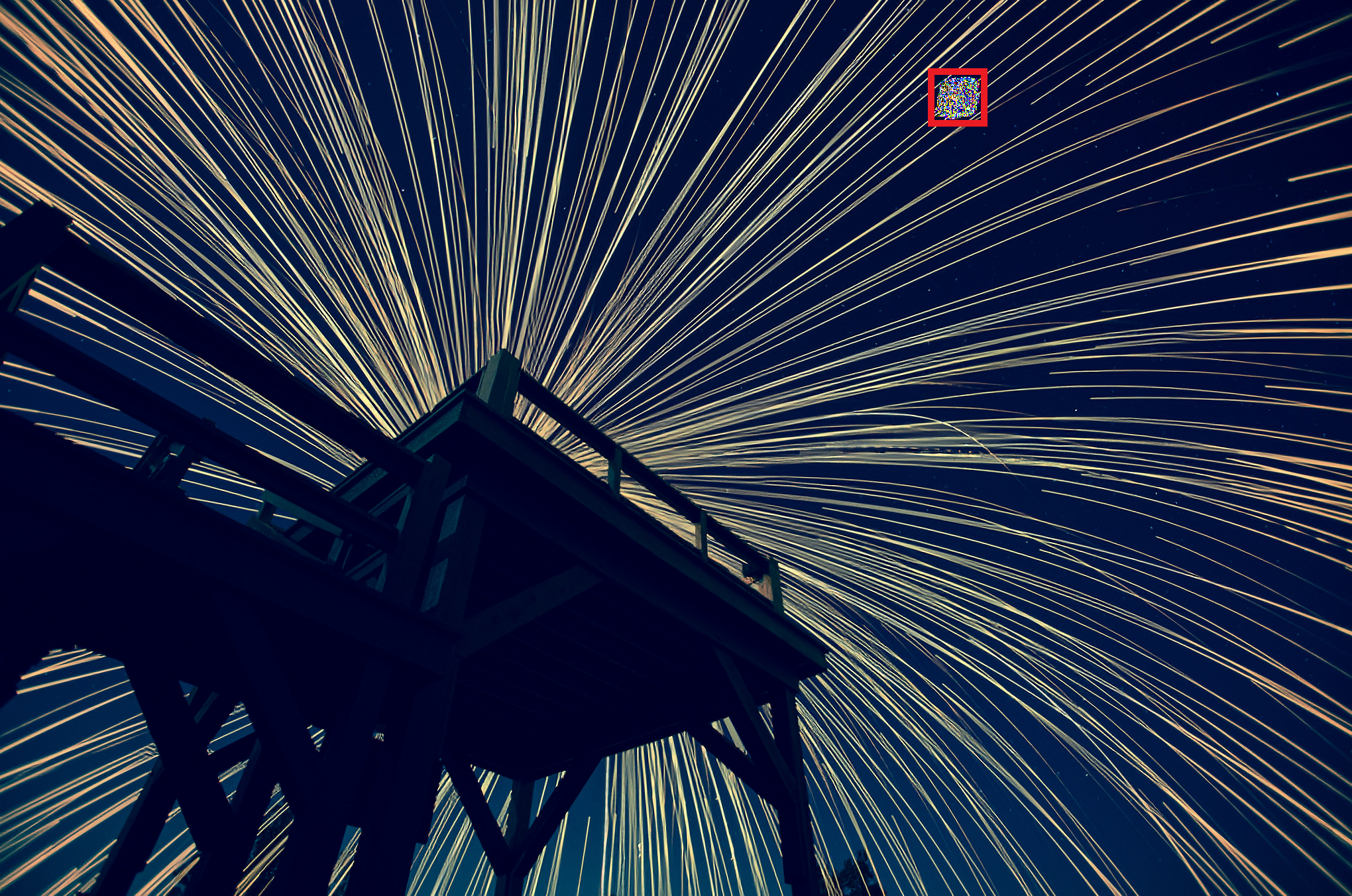}
    \caption{Failure case of FS-NCSR}
    \end{subfigure}
    
    \begin{subfigure}{.45\textwidth}
    \centering
    \includegraphics[width=.95\linewidth]{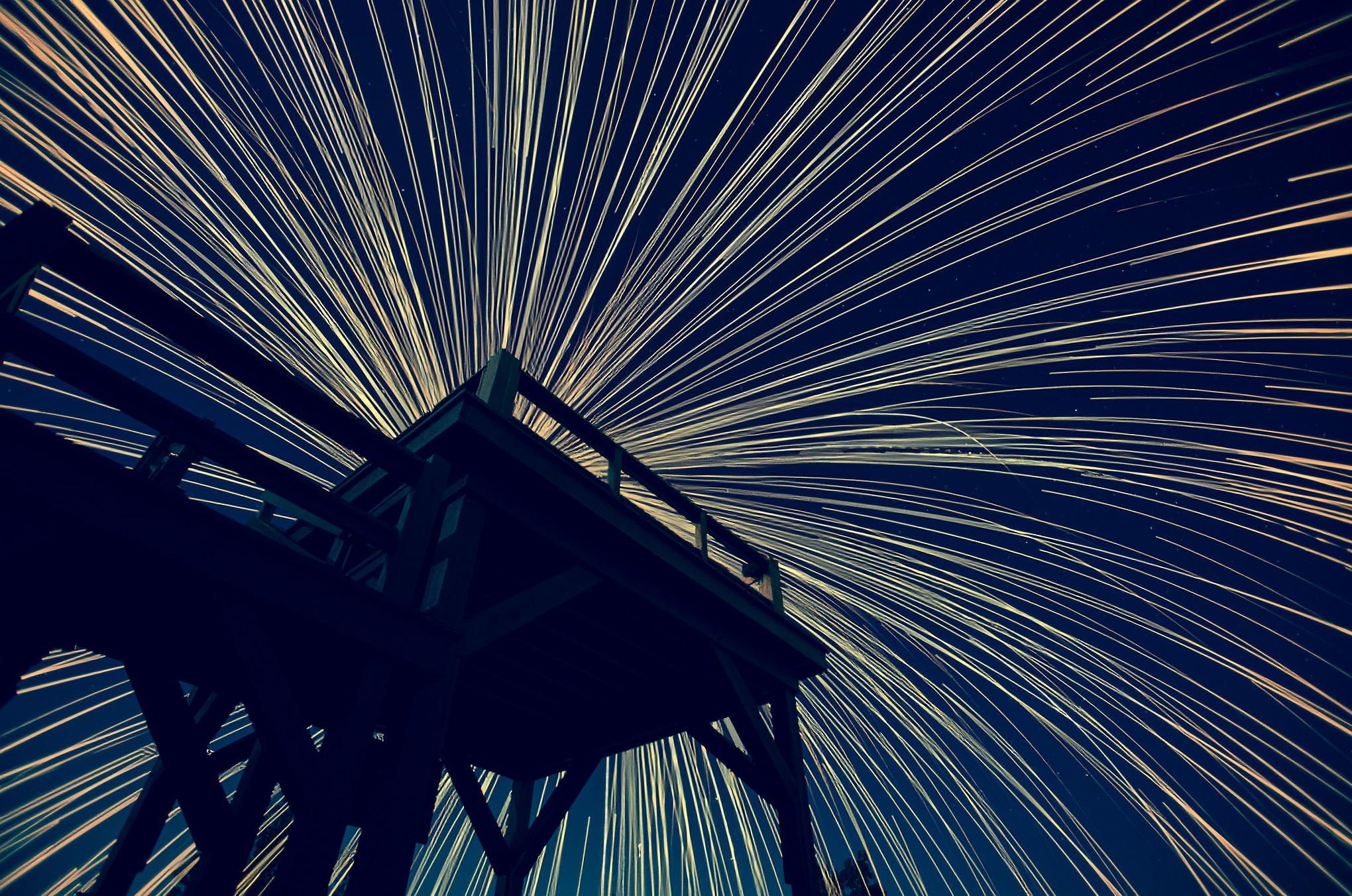}
    \caption{FS-NCSR Ouput without any artifact}
    \end{subfigure}
    \begin{subfigure}{.45\textwidth}
    \centering
    \includegraphics[width=.95\linewidth]{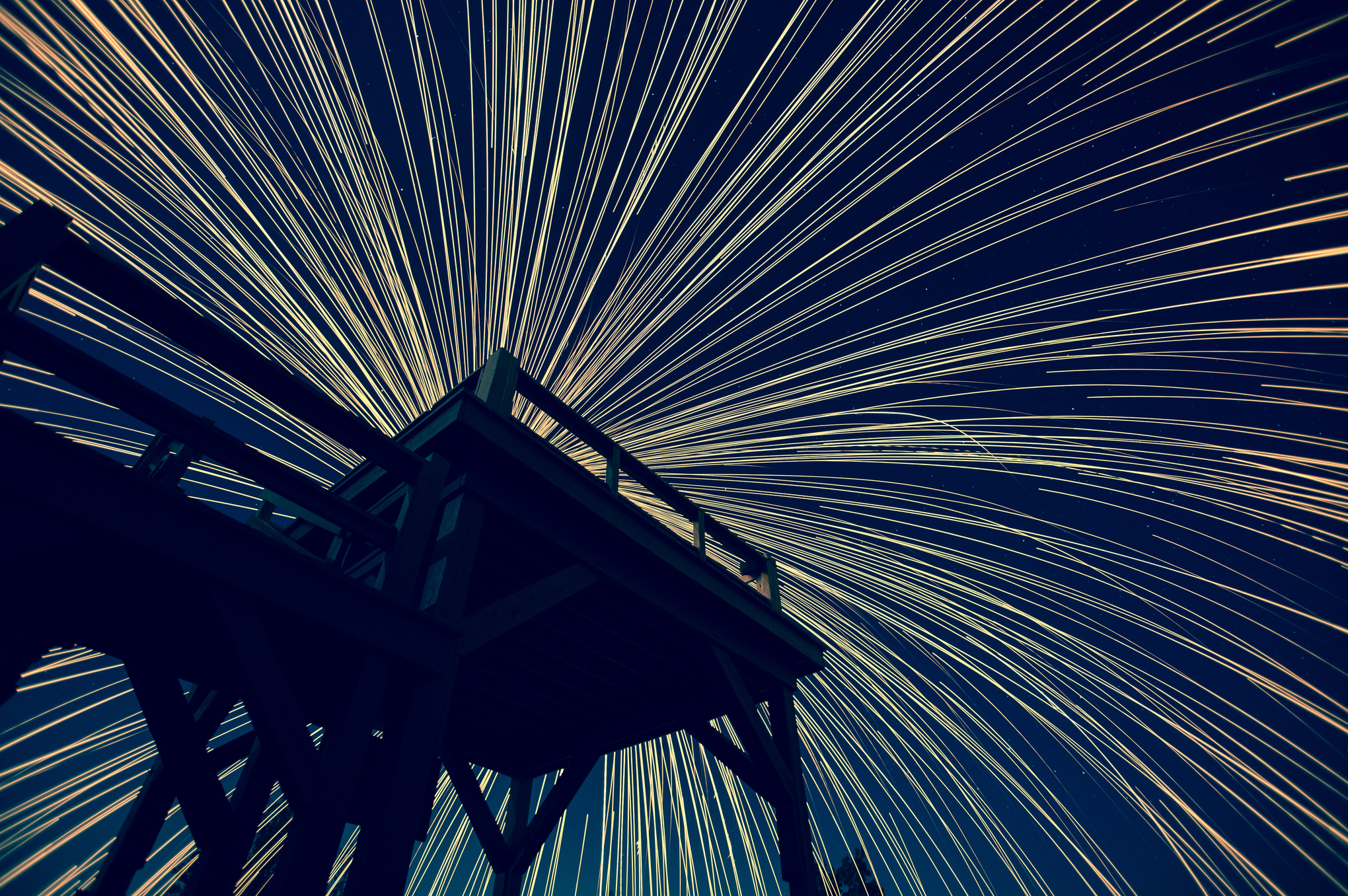}
    \caption{Ground Truth: 0828 from DIV2K}
    \end{subfigure}
\caption{Visual comparison of failure cases on SR $\times$4 results from NCSR and FS-NCSR. The degree of the artifact is relatively less in the result of FS-NCSR than that of NCSR. And FS-NCSR could generate clear SR output without any artifact while NCSR couldn't. Each output was chosen randomly.}
\label{fig:artifact}
\end{center}
\end{figure*}

\subsection{Qualitative Results} \label{sub:qualitative}

The qualitative result in Figure \ref{fig:diversity} shows that the direction and the degree of density of the leaves are slightly different for every 5 outputs. 
Thus, we can say that the proposed method not only shows a higher diversity score than previous approaches but also can generate outputs with diverse details that are distinguishable visually. It means that the frequency separation can enhance the high mode coverage performance of the flow-based model.

We now qualitatively compare our result with the output of NCSR to verify the effect of the frequency separation. As discussed in \ref{sub:quantitative}, the FS-NCSR's LPIPS was lower than the existing approaches. But Figure \ref{fig:compare_crop} shows that FS-NCSR can reproduce the characters more clearly than NCSR. This qualitatively confirmed that although the training focused on high-frequency information performs slightly lower on LPIPS, actual outputs do not suffer a degradation of image quality than the existing methodologies.

The existing SRFlow and NCSR models show repeated failure cases where artifacts appear in a specific image (\eg 0807, 0828 from DIV2K validation set). In the case of 0807 from DIV2K, for instance, when both SRFlow and NCSR generated the corresponding $\times$4 super-resolved outputs, all outputs were failure cases since some artifacts appeared. 
On the other hand, when FS-NCSR generate the $\times$4 super-resolved outputs of the given image, 4 out of 10 outputs were made without any artifact, and even for 6 failure cases, the degree of the artifact was relatively less than that of NCSR. Figure \ref{fig:artifact} presents the degree of the artifact differs between NCSR and FS-NCSR output and the FS-NCSR's artifact-free results compared to the ground truth image.

\subsection{Ablation: Comparison of Generated High-Frequency Information}

So far, we have discussed the results both quantitatively and qualitatively with the super-resolved outputs only. But we tried to compare the results from the perspective of frequency information additionally. 
Since the sparse high-frequency information plays a key role in the proposed method, we investigated how the proposed method affects the sparsity of the generated high-frequency information.
For this purpose, the generated high-frequency information in $[0,1]$ range is first quantized to the uint8 $[0,255]$ range. And then \textbf{Sparsity} and \textbf{Relative Sparsity (RS)} is computed as follows:
\begin{equation}\label{eq:sparsity}
\begin{aligned}
    \textbf{Sparsity} = 1 - \frac{\textbf{Number of non-zero pixels}}{H\times W\times C} \\
    \textbf{RS} = 1 - \frac{\textbf{Number of non-zero pixels}}{\textbf{Number of non-zero gt pixels}}
\end{aligned}
\end{equation}
where $(H, W, C)$ is the shape of a given image.
Since the ground truth high-frequency information is already sparse, the RS reflects the sparsity of each ground truth image for a more fair comparison.

\begin{table}[ht]
\begin{tabular}{@{}lll@{}}
\toprule
Model & Average Sparsity & Average RS \\ \midrule
NCSR~\cite{kim2021noise}  & 66.2\% & 1.123   \\
\textbf{FS-NCSR (Ours)} & \textbf{66.0\%} & \textbf{1.120}  \\ 
\bottomrule
\end{tabular}
\captionsetup{justification=raggedright,singlelinecheck=false}
\vspace{-0.3cm}
\caption{As a $\times$4 super-resolution results on DIV2K validation set, \textbf{Average Sparsity of GT high frequency information is 59.0\%}.}
\label{tab:Freq_space}
\end{table}

See Table \ref{tab:Freq_space}. Although our proposed method shows less average sparsity and average RS slightly, the average sparsity of the ground truth high-frequency information and the that of generated output from both NCSR and FS-NCSR was about 10\% more sparse, resulting in a lack of information compared to the ground truth.
This margin of difference with the ground truth verifies that a significant loss of information still exists from the perspective of the frequency domain. Therefore, it seems that it needs to be addressed in future studies.

%% file: paper/5_ntire.tex
\begin{table}[t]
\scalebox{0.9}{
\begin{tabular}{@{}lllll@{}}
\toprule
Team & LPIPS & LR PSNR & Div. Score & MOR \\ 
\midrule
IMAG\textunderscore ZW & 0.171 & 48.14 & 21.938 & 3.57  \\
\textbf{FS-NCSR (Ours)} & 0.126 & 50.13 & \textbf{28.853} & 3.67 \\ 
IMAG\textunderscore WZ & 0.169 & 45.20 & 27.320 & \textbf{3.34}  \\
\midrule
SSS & 0.110 & 44.70 & 13.285 & \textunderscore  \\
NCSR & 0.117 & 50.54 & 26.041 & \textunderscore  \\
SRFlow & 0.122 & 49.86 & 25.008 & 3.62  \\
ESRGAN & 0.124 & 38.74 & 0.000 & 3.52 \\
\bottomrule
\end{tabular}
}
\captionsetup{justification=raggedright,singlelinecheck=false}
\vspace{-0.3cm}
\caption{Quantitative results for NTIRE 2022 Challenge
on "Learning Super Resolution Space" on $\times$4 track. The results were taken from \cite{lugmayr2022ntire}. The top block of the table is this year's result.}
\label{tab:4x_result}
\end{table}

\begin{table}[t]
\scalebox{0.9}{
\begin{tabular}{@{}lllll@{}}
\toprule
Team & LPIPS & LR PSNR & Div. Score & MOR \\ 
\midrule
\textbf{FS-NCSR (Ours)} & 0.257 & 50.37 & 26.539 & 4.510 \\ 
\midrule
SSS & 0.237 & 37.43 & 13.548 & 4.850  \\
NCSR & 0.259 & 48.64 & 26.941 & 4.503  \\
SRFlow & 0.282 & 47.72 & 25.582 & 4.775  \\
ESRGAN & 0.284 & 30.65 & 0.000 & 4.452 \\
\bottomrule
\end{tabular}
}
\captionsetup{justification=raggedright,singlelinecheck=false}
\vspace{-0.3cm}
\caption{Quantitative results for NTIRE 2022 Challenge
on "Learning Super Resolution Space" on $\times$8 track. The results were taken from \cite{lugmayr2022ntire}. The top block of the table is this year's result.}
\label{tab:8x_result}
\end{table}

Our proposed method, FS-NCSR, achieved competitive results in both tracks of NTIRE 2022 "Learning Super Resolution Space Challenge" \cite{lugmayr2022ntire}. See table \ref{tab:4x_result} and \ref{tab:8x_result} for the challenge result of $\times$4 and $\times$8 tracks respectively. 
In the $\times$4 track, FS-NCSR obtained the highest diversity score among the existing and newly proposed methods by a relatively large margin. Also, it obtained the best LPIPS and LR-PSNR results among this year's participants, although it did not lead to the best MOR.
In the $\times$8 track, FS-NCSR was this year's only method that achieved comparable results compared to the last year's approaches. Through the improvement of LR-PSNR, it seems that the frequency separation affected improving the consistency with low-resolution.

%% file: paper/6_conclusion.tex
We propose a flow-based algorithm, FC-NCSR, to learn high-frequency information of super-resolution space. 
Based on the relation between the high-frequency information and the high-quality details of the given image, we train the generative model for super-resolution to produce the high-frequency information corresponding to the low-resolution input.
With a simple high-pass filter using the low-frequency information of the low-resolution input, we successfully increase the super-resolution diversity without any influence on the stability of the flow-based NLL training and visual quality degradation.
We also confirm that the frequency separation of FS-NCSR reduces the failure cases due to artifacts, and therefore, significantly improves the quality of the super-resolution output.